\documentclass{article}

\usepackage[final, nonatbib]{neurips_2023_ml4ps}
\usepackage[pdftex]{graphicx}
\usepackage{subcaption}
\usepackage[utf8]{inputenc} 
\usepackage[T1]{fontenc}    
\usepackage{hyperref}       
\usepackage{url}            
\usepackage{booktabs}       
\usepackage{amsfonts}       
\usepackage{nicefrac}       
\usepackage{microtype}      
\usepackage{xcolor}         

\title{Ultra Fast Transformers on FPGAs for Particle Physics Experiments}

\author{%
    Zhixing Jiang \\
    University of Washington \\
    \texttt{Zhixij@uw.edu} \\
    \AND
    Dennis Yin \\
    University of Washington \\
    \texttt{lostecho@uw.edu} \\
    \And
    Elham E Khoda \\
    University of Washington \\
    \texttt{ekhoda@uw.edu} \\
    \And
    Vladimir Loncar\thanks{Also at Institute of Physics Belgrade, Serbia} \\
    Massachusetts Institute of Technology \\
    \texttt{vloncar@mit.edu} \\
    \And
    Ekaterina Govorkova \\
    Massachusetts Institute of Technology \\
    \texttt{katyag@mit.edu} \\
    \And
    Eric Moreno \\
    Massachusetts Institute of Technology \\
    \texttt{Emoreno@mit.edu} \\
    \And
    Philip Harris \\
    Massachusetts Institute of Technology \\
    \texttt{pcharris@mit.edu} \\
    \And
    Scott Hauck \\
    University of Washington \\
    \texttt{hauck@uw.edu} \\
    \And
    Shih-Chieh Hsu \\
    University of Washington \\
    \texttt{schsu@uw.edu} \\
}

\newcommand{\pt}{p_{\mathrm{T}}}

\begin{document}

\maketitle

\begin{abstract}
This work introduces a highly efficient implementation of the transformer architecture on a Field-Programmable Gate Array (FPGA) by using the \texttt{hls4ml} tool. 
Given the demonstrated effectiveness of transformer models in addressing a wide range of problems, their application in experimental triggers within particle physics becomes a subject of significant interest.
In this work, we have implemented critical components of a transformer model, such as multi-head attention and softmax layers.
To evaluate the effectiveness of our implementation, we have focused on a particle physics jet flavor tagging problem, employing a public dataset.
We recorded latency under 2 $\mu$s on the Xilinx UltraScale+ FPGA, which is compatible with hardware trigger requirements at the CERN Large Hadron Collider experiments.

\end{abstract}

\section{Introduction}

Accelerated Machine Learning (ML) inference is necessary to run the algorithms in the online event selection systems of the particle physics experiments.
Due to the extremely high particle collision frequency of 40 MHz at the Large Hadron Collider (LHC) \cite{LHC_2008} at CERN \cite{CERN_as}, it is impossible to read out and store all the collision events.
As a result, the LHC experiments \cite{ATLAS_2008, ALICE_2008, CMS_2008, LHCb_2008},  try to read out only the interesting via an online selection system called the trigger. 
Most of the LHC experiments use a two-stage trigger system, hardware-based Level-1 trigger and software-based High-Level trigger.
The Level-1 trigger operates at 40 MHz, so the algorithms usually run on application-specific integrated circuits (ASICs) or FPGAs.
As the average number of collisions at the LHC is expected to increase with time, sophisticated ML algorithms will be crucial for Level-1 triggers to efficiently filter events.
There have been numerous efforts to port ML algorithms like Deep Neural Networks \cite{Duarte_2018}, Convolution Neural Networks \cite{CNN_hls4ml}, Recursive Neural Networks \cite{RNN_hls4ml, RNN_GW_hls4ml}, Graph Neural Networks \cite{GNN_hls4ml} onto FPGAs for physics applications using High-Level Synthesis (HLS) languages with the \texttt{hls4ml} package \cite{Duarte_2018, Fahim:2021cic}. \texttt{hls4ml} is an HLS-based compiler for a neural network to FPGA firmware conversion.

In recent years, the transformer \cite{vaswani2023attention} architecture became popular for their great performance in language modeling tasks like encoder-only BERT \cite{bert}, decoder-only GPT \cite{gpt}, etc.
Over time, the utility of transformer models extended beyond language modeling, impacting a wide range of ML applications.
They are now widely used in particle physics for offline computing tasks like particle reconstruction \cite{Shmakov:2021qdz}, identification \cite{Qu:2022mxj, ABCNet, Mikuni_2021}, etc.
Often the transformer-based models show better performance over other architecture, but they are very compute intensive and suffer from a slow inference rate.
Because of the computationally intensive nature, it becomes challenging to implement \cite{li2020ftrans, 9424344, 9976354, hong2022dfx} them on hardware like FPGAs, where a limited amount of resources is available. 
Another previous work \cite{imperial_transformer} explored this design space in the context of a particle physics experiment by studying a small transformer for jet classification.

In this work, we present a flexible and efficient implementation of transformers written in HLS for the \texttt{hls4ml} package.
This integration into \texttt{hls4ml} opens the door for wider low-latency applications of the Transformer models.
Here the main focus is on the trigger applications in the LHC experiments. 
However, our implementation is very general and it is relevant to many real-time detector systems across fundamental science where low-latency high throughput inference is necessary.

\section{Benchmark study}
\label{benchmark}

To benchmark our implementations, we study the open data samples from the Compact Muon Solenoid (CMS) experiment which contain top quark pairs decaying hadronically with center-of-mass energy of 7 TeV \cite{ftag_dataset}. 
These events contain many bottom quark jets (b jets), charm quark jets (c jets) and jets from light quarks, and gluons (light jets) originating from top quark decay.
Jets are collimated showers of particles that result from the decay and hadronization of quarks q and gluons g. 
At the LHC, an interesting jet signature emerges from overlapping quark-initiated showers produced in decays of heavy standard model particles like bottom quarks.
The jets in the dataset are labeled as b, c, and light jets depending on whether they contain bottom quarks,  charm quarks, or neither, respectively.
The task of identifying the heavy jets like b jets from c jets and light jets is called flavor tagging. 
The main feature that separates b jets (and c jets) from light jets is the presence of the displaced vertex corresponding to the decay of the hadron containing the b (or c) quark. 
These hadrons are long-lived due to their mass, and the decay time depends on their momenta. 
Our proposed algorithm aims to identify the presence of tracks that are consistent with these displaced vertices using a transformer architecture. 

All the jets are reconstructed using the anti-kt algorithm with a distance parameter of $R = 0.5$.
The jets are required to have transverse momenta ($\pt$) larger than 30 GeV and absolute pseudorapidities less than 2.0. 
Charged particle tracks with $\pt$ larger than 1 GeV are associated with the nearest jet if they are within the angular distance $\Delta R$ (track, jet) of 0.5.
Tracks within a jet are ordered by the significance of their transverse impact parameter ($\mathcal{S}(d_0)$), and only the first 15 tracks are used for this study. 
Each track is represented by a vector of six features: transverse and longitudinal impact parameters ($d_0$, $d_z$) and their significances ($\mathcal{S}(d_0)$, $\mathcal{S}(d_z)$),  $\Delta R(\mathrm{track, jet})$, and relative transverse momentum between the track and the jet ($\pt(\mathrm{track})/\pt(\mathrm{jet})$).

The flavor tagging classifier model is constructed using Keras+TensorFlow, using a transformer architecture with 9135 trainable parameters. 
The padded sequence of tracks, with a maximum length of 15, is directly fed into a transformer encoder block. 
No positional encoding is used, as the ordering is not crucial for this problem.
Each encoder block contains a multi-head attention (MHA) layer with two heads, running two scaled dot-product attention layers in parallel, and a feed forward network with two dense layers.
Outputs of the MHA layer are passed through a feed forward block where the layer dimensions are 8 and 6, respectively.
Due to the simplicity of the flavor tagging problem, we did not include a layer normalization after the MHA layer. 
The structure of the encoder block is shown in Fig. \ref{fig:transformer_block}.
The outputs of the encoder blocks are flattened and passed through three dense layers with 32, 16, and 8 units. 
The output layer uses a softmax function and predicts three class probabilities corresponding to b, c, and light jets.
The model contains three encoder blocks and the architecture is shown in Fig. \ref{fig:model}.
The training is performed with a categorical cross-entropy loss, with 30\% of the training data retained for validation and testing.

\section{Implementations}

\begin{figure*}[!ht]
    \begin{subfigure}[b]{0.33\linewidth}
     \centering
        \includegraphics[scale=0.5]{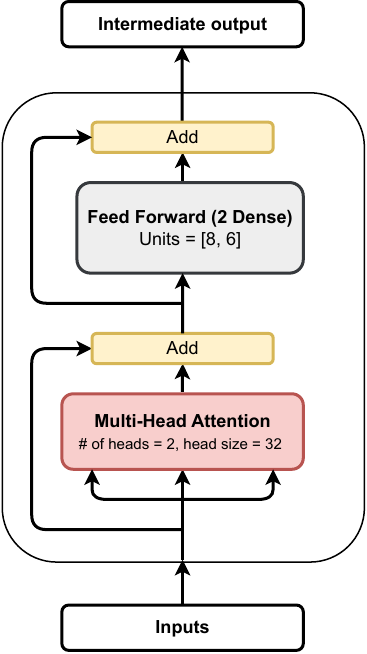}
        \caption{Transformer encoder block}
        \label{fig:transformer_block}
    \end{subfigure}
    \begin{subfigure}[c]{0.28\linewidth}
    \centering
    \vspace{-5.7cm}
        \includegraphics[scale=0.6]{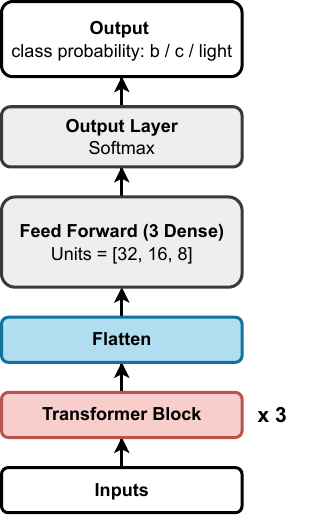}
        \caption{Model architecture}
        \label{fig:model}
    \end{subfigure}
    \begin{subfigure}[c]{0.28\linewidth}
    \vspace{-6cm}
    \centering
        \includegraphics[scale=0.5]{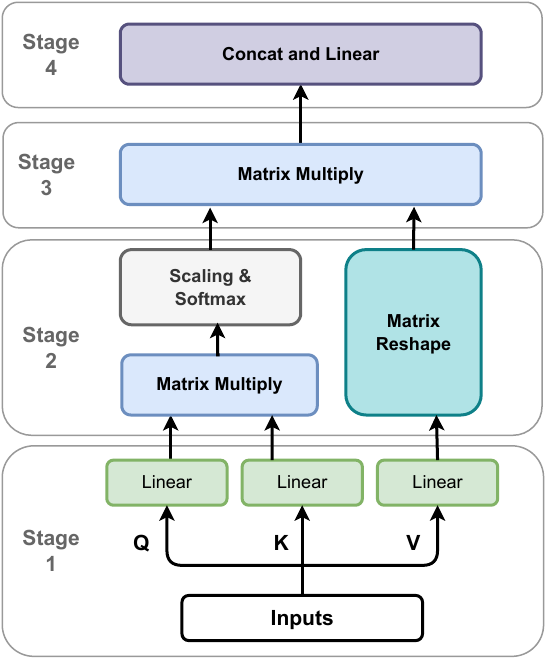}
        \caption{Pipeline stages}
        \label{fig:mha_stages}
    \end{subfigure}
    \caption{The encoder block used for the transformer model is shown in (a). The full model architecture is shown in (b). The pipeline stages for the multi-head attention layer is shown in (c).}
    
\end{figure*}

One of the main focuses of this work is to implement the MHA layer in HLS.
The implementation of the MHA layer is divided into four sequential pipeline stages shown in Fig.~\ref{fig:mha_stages}.
We optimized the performance to increase throughput by pipelining the different stages of MHA as described below:

\underline
{The first stage} entails a linear projection transforming input into Query (Q), Key (K), and Value (V) vectors via distinct weight matrices. At each step, matrix-vector multiplication is pipelined for efficiency. 
Vectors required once are stored in FIFO memory, while frequently accessed data reside in BRAM blocks. Continuous DSP engagement ensures maximal computational unit usage. 
This method complements sequential data processing by optimizing memory use and promoting efficient data flow for later stages. Bandwidth is enhanced by stacking multiple FIFOs.

\underline{The second stage} starts computing the attention mechanism by taking the dot product of the Q  and K vectors, producing a relevance score for each element of the input sequence.
The product is then divided by the dimension of the key vectors, \(\sqrt{d_k}\), before passing it through a lookup table-based softmax function.
The softmax has two lookup tables (LUT), one for inverse function, and one for exponential function. LUTs are used to reduce the computation intensity. 
Additionally, V matrix undergoes reshaping to allow simultaneous row-wise and column-wise access, crucial for Stage 3. Efficient FPGA implementation requires orchestrated data flow and resource allocation. Hence, 
K vectors are preloaded into a 2D register for parallel retrieval, essential for row-by-row matrix multiplication.

\underline{The third stage} involves the matrix multiplication of the scores matrix and the corresponding V vectors. 
The V vectors are stored in a fully accessible register for the parallel multiplication process.
The results are stored back in the FIFO memory and passed to the next stage of processing. 

\underline{The fourth stage} includes two key processes: the concatenation of the output from all attention heads and the subsequent linear transformation of the concatenated result. 
Each attention head provides an output vector loaded row by row, aligning with the temporal sequencing of the data. 
Once loaded, the outputs are concatenated together to form a single, unified data stream. 
Then the data stream is passed through a linear layer. 
The linear layer is also pipelined, and it inputs and outputs one row of data at a time.
This stage manages the output from all heads and efficiently generates the final output.

\vspace{0.5cm}

Apart from implementing the MHA layer, we have also optimized the softmax HLS implementation inside the \texttt{hls4ml} tool to reduce the computational cost. Softmax is used many times in the model, so it is crucial to have an efficient HLS implementation to run inference on an FPGA.

\section{Results}
The flavor tagging model described in Sec. \ref{benchmark} is translated into an HLS model using the \texttt{hls4ml} framework.
Our tests were done using Vivado HLS 2020.1 with a Xilinx UltraScale+ FPGA VU13P (part number \texttt{xcvu13p-fhga2104-2L-e}) as the target device.
For the HLS implementation two different optimizations are studied: quantization and parallelization. 

\begin{figure}[!ht] 
    \centering
    \begin{subfigure}[c]{0.325\linewidth}
    \centering
        \includegraphics[width=1\textwidth]{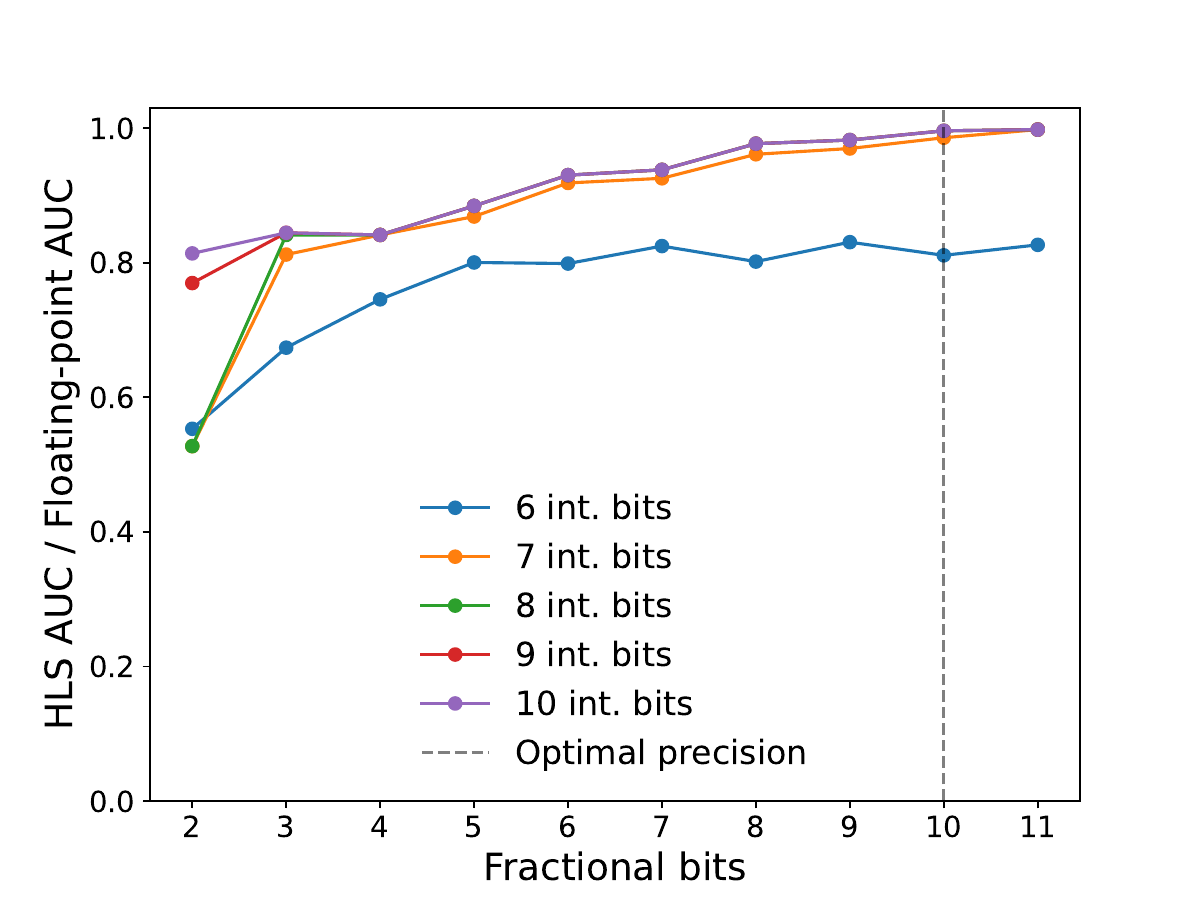}
        \caption{AUC ratio}
        \label{fig:auc}
    \end{subfigure}
    \begin{subfigure}[c]{0.325\linewidth}
    \centering
        \includegraphics[width=1\textwidth]{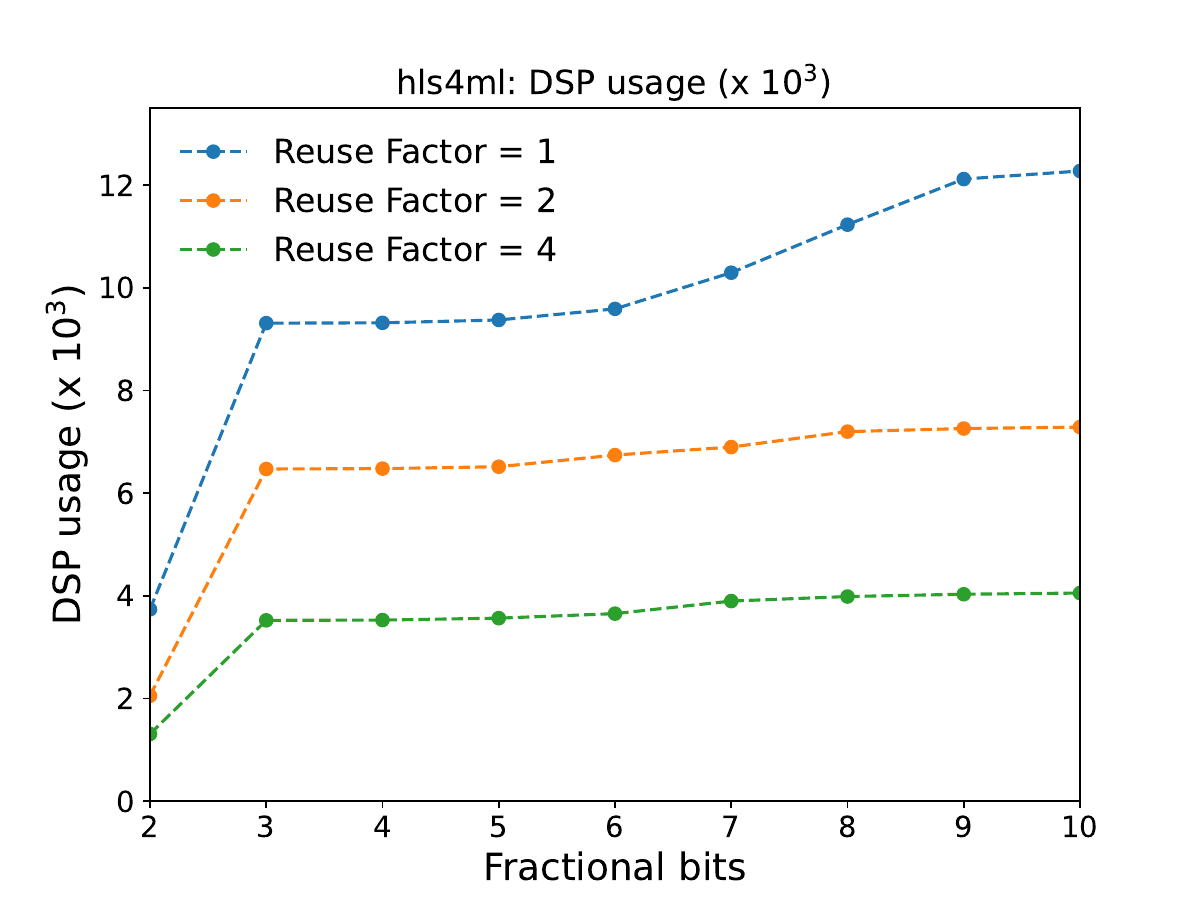}
        \caption{DSP usage}
        \label{fig:dsp_util}
    \end{subfigure}
    \begin{subfigure}[c]{0.325\linewidth}
    \centering
        \includegraphics[width=1\textwidth]{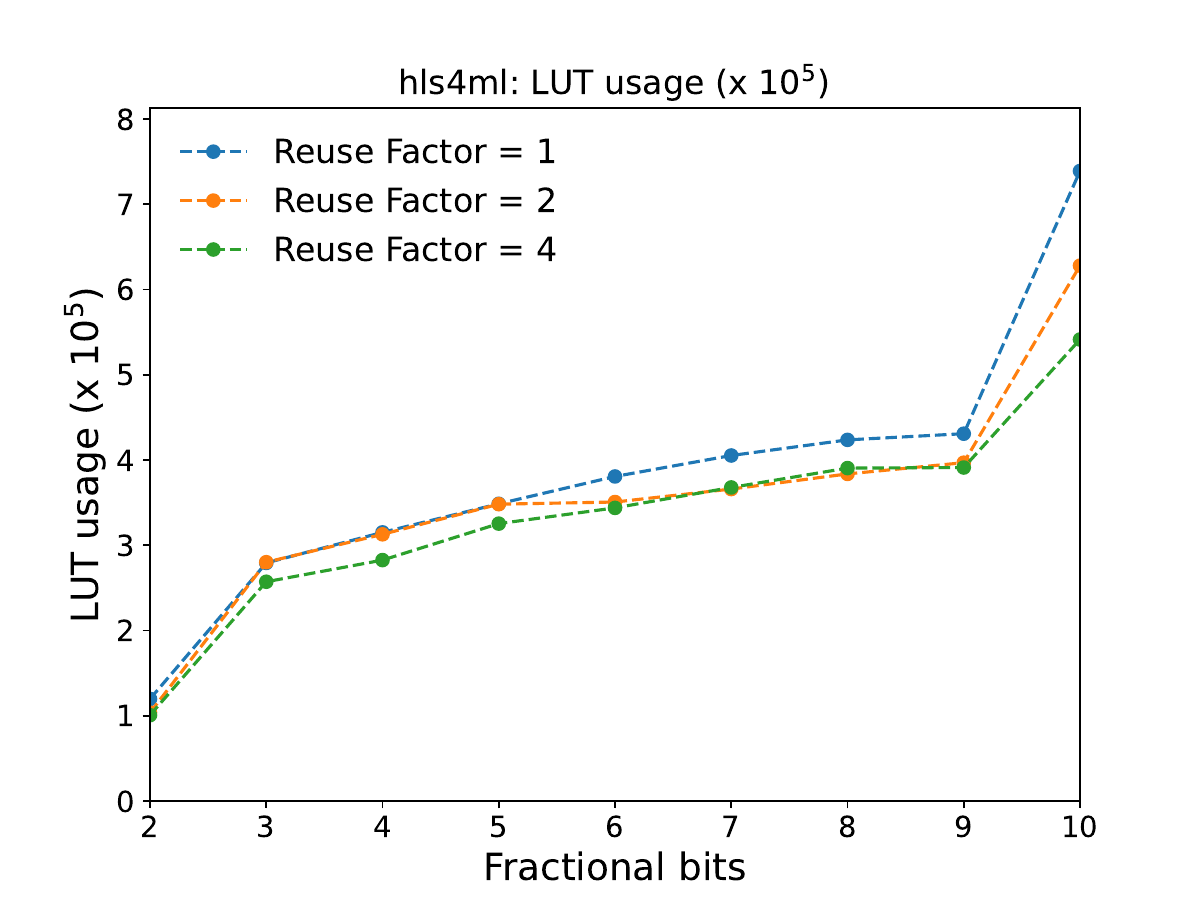}
        \caption{LUT usage}
        \label{fig:lut_util}
    \end{subfigure}
    \caption{ (a) Ratios of the fixed-point and floating-point AUCs as function of fractional bits. Five different values between 6 and 10 bits are chosen for the integer precision. Utilization of (b) DSP and (c) Lookup tables are shown as a function of fractional bits while keeping the integer part fixed to 10. Three different configurations with reuse factor of 1 (blue), 2 (orange), or 4 (green) are shown. The target board (part number \texttt{xcvu13p-fhga2104-2L-e}) has a total of 12288 DSPs and 1.72 million LUTs. }
    
\end{figure}

The quantization process reduces the numerical precision of the model parameters, such as weights and biases, as well as inputs.
Typically, ML model parameters are stored as 32-bit floating-point numbers
Although floating-point numbers offer an extensive dynamic range, they consume significant computing resources when implemented on an FPGA. 
Therefore, for FPGA implementation, fixed-point numbers with fixed precision are preferred. 
This shift to fixed-point representation greatly accelerates computation by reducing both computational resource usage and memory utilization.
In our study, we systematically explore fractional bit variations while maintaining a fixed precision of 6, 7, 8, 9, or 10 bits for the integer part. 
We evaluate the receiver operating characteristic (ROC) curve for the transformer-based classifier employing the area under the curve (AUC) as a performance metric.
The ratio of the AUCs (fixed-point HLS model / floating-point Keras model) is shown in Fig. \ref{fig:auc} as a function of fraction bits for integer bits.
From the figure, it is clear that we need at least 10 integer bits and 10 fractional bits to get a similar performance as the floating-point model.

The \texttt{hls4ml} offers a valuable feature known as the ``reuse factor" parameter, which plays a pivotal role in governing the optimization of parallelization and the efficient utilization of computing resources. 
This factor determines the number of times each multiplier is used for computing the neuron values within a given layer. 
If the reuse factor is set to 1, the computation becomes fully parallel,  as each multiplication operation is executed independently by a dedicated digital signal processing (DSP) block.
As we increase the reuse factor the computing resource utilization decreases, but the latency increases proportionally. 
To study the resource-latency trade-off and find an optimal implementation for our model, we have synthesized (full Vivado synthesis) it with varying values of the reuse factor and fractional bit precision.
For each case, we quantified the utilization of FPGA resources of different categories like memory (BRAM), DSPs, flip-flops (FFs), and lookup tables (LUTs).
The utilization of DSPs and LUTs are shown in Fig. \ref{fig:dsp_util} and Fig. \ref{fig:lut_util}, respectively, as a function of fractional bits (integer bit = 10) for reuse factors of 1, 2, or 4.
As anticipated, the resource utilization goes up as we reduce the reuse factor. 
It's worth noting that the target board has a total of 12288 DSPs and 1.72 million LUTs, providing us with flexibility in selecting any of the three reuse factors to achieve the optimal precision of (int. = 10, frac. = 10) during the model synthesis.

Remarkably, the observed latency aligns with the requirements of the LHC hardware trigger.
For the fully parallel scenario with a reuse factor 1, the model's inference latency is 2.077 $\mu s$. 
Here, the clock period is 6.58 ns, resulting in output generation every 49 clock cycles or 322.42 ns.
However, the latency increases to 3.467 $\mu s$ and 5.853 $\mu s$ for reuse factors of 2 and 4, respectively.

\section{Summary and Outlook}

We have successfully implemented a transformer architecture with multi-head attention in HLS for FPGA inference.
This implementation has been seamlessly incorporated into the \texttt{hls4ml} package, which facilitates the automatic translation of transformer models for low-latency inference applications.
It is essential to note that some critical features, including positional encoding and layer normalization, have been left for future work.
To demonstrate the effectiveness of the current implementation, we conducted a  study using a flavor tagging model.
Notably, the model's inference latency falls within a range of 2 to 6 $\mu s$, fully complying with the stringent timing constraints of the hardware triggers. 
What sets our implementation apart is its exceptional versatility.
It can readily adapt to models with different configurations, such as varying sequence lengths and the number of attention heads, developed using Keras and TensorFlow.
As a result, this integration signifies a significant development and opens avenues for the broader adoption of low-latency applications using transformer models.

\section{Broader Impact}

Although we demonstrate the performance of one specific algorithm here, this work could be used to accelerate other reconstruction algorithms in particle physics experiments. 
In fact, \texttt{hls4ml} transformer can be used for low latency inference for other scientific domains like neuroscience, gravitational wave, material science, etc., and various non-scientific domains.

\section*{Acknowledgments}
We acknowledge the Fast Machine Learning collective and the NSF A3D3 as open communities of multi-domain experts and collaborators.
These communities were important for the development of this project.
This research was funded in part by National Science Foundation (NSF) grants No. 1934360, 1934700,  2117997, 1934360, and 1931469.

\section*{Code Availability Statement}

The \texttt{hls4ml} library is available at \href{https://github.com/fastmachinelearning/hls4ml}{https://github.com/fastmachinelearning/hls4ml} and MHA support will be made available in the near future. 
For examples of how to use \texttt{hls4ml}, the notebooks in
\href{https://github.com/fastmachinelearning/hls4ml-tutorial}{https://github.com/fastmachinelearning/hls4ml-tutorial} serve as a general introduction.


\bibliography{reference}

\end{document}